# On Safety Assessment of Artificial Intelligence

Jens Braband, Siemens Mobility GmbH

Hendrik Schäbe, TÜV Rheinland

## Abstract

In this paper we discuss how systems with Artificial Intelligence (AI) can undergo safety assessment. This is relevant, if AI is used in safety related applications. Taking a deeper look into AI models, we show, that many models of artificial intelligence, in particular machine learning, are statistical models. Safety assessment would then have to concentrate on the model that is used in AI, besides the normal assessment procedure. Part of the budget of dangerous random failures for the relevant safety integrity level needs to be used for the probabilistic faulty behavior of the AI system. We demonstrate our thoughts with a simple example and propose a research challenge that may be decisive for the use of AI in safety-related systems.

## Introduction

In the last years, artificial intelligence (AI) has become more and more popular and an increasing number of applications has been reported. These include for example

- Data processing

- Assistance systems

- Speech recognition

- Face recognition

- Nursing robots

- Autonomous driving systems

- Art etc.

Some of the applications of artificial intelligence may be safety relevant. Then, functional safety standards should be applied and as a consequence, safety assessment is required.

In this paper, we consider safety assessment of systems with AI. In the second section we describe, what AI means. In the third section we show, how a safety integrity level for AI systems can be obtained. In section four we will take a deeper view into AI systems – this is necessary to understand AI systems and to have an approach to them in terms of functional safety. In the fifth section, we describe the requirements of the functional safety standards for AI systems and a possible assessment procedure. In section six, we provide an example, of how safety assessment could be carried out on a very simple system. In the last section, we present our conclusions.

## What is artificial intelligence?

There exist many publications and many systems are named as being artificially intelligent. An overview can be found e.g. in Brunette et al (2009). The starting point has been the Turing test in the 50s, which is intended to check whether a computer exhibits intelligent behavior, comparable to that of a human being. Later on, the concept of evolutionary programs has been established. The term „Artificial Intelligence" has first been used at Dartmouth College in 1956. In the meanwhile, different concepts have been proposed by many researchers.

Artificial Intelligence can be defined as intelligence demonstrated by machines. Artificial intelligence mimics cognitive functions, learning, problem solving etc.

A question is, whether the following are criteria of intelligence points would be criteria for artificial intelligence or not:

- use of speech,

- consciousness,

- self-awareness.

But while there are truly astounding results, there are many articles and presentations about the „deep learning hype", see e.g. Hättasch&Geisler (2019), and as far as we know there is so far no published complete safety argument for any AI application, but there are many research projects on safety justifications for AI.

However some approaches have been recently made from a safety point of view, most notably the draft UL 4600 standard (2019), which demands a safety case approach for autonomous vehicles, that may utilize AI algorithms. However also UL 4600 elaborates only on What to argue, but not the How. This is clearly described in the preface: "Conformance with this standard is not a guarantee of a safe automated vehicle." Its emphasis is rather on "repeatable assessment of the thoroughness of a safety case". UL 4600 is intended be used as an extension of IEC 61508.

Other standardization committees, e. g. the German DKE, focus on a process and lifecycle oriented approach. Putzer (2019) propagates a $\lambda_{AI}$, a measure similar to a hazard rate in functional safety, but gives no concise definition.

## Does AI need a SIL?

In this section we will discuss, whether we would need a safety integrity level for artificial intelligence and if yes, how it should be determined.

The concept of Safety Integrity Level (SIL) is used in many standards for functional safety. The mother standard is the well-known IEC 61508. The reader may be referred to Schäbe (2018) for the determination of SILs.

The following figure 1 shows the situation with a normal electric, electronic, programmable electronic system (E/E/PE system). Here, we have an equipment under control, information form sensors that enter the control system and actors operated by the control system. Depending on the consequences of faulty behavior of the control system, the latter gets a safety integrity level (SIL).

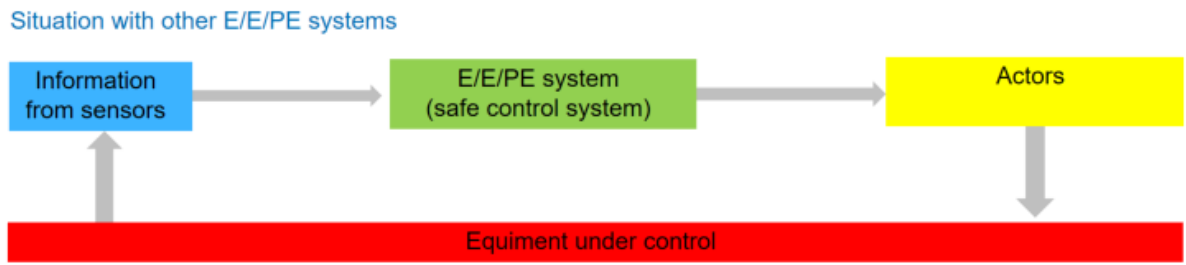

Figure 1        E/E/PE Control system

Now, it does not matter what type of control system we have. For the hazard analysis and the determination of the SIL it is considered as a black box anyway. This is depicted in figure 2.

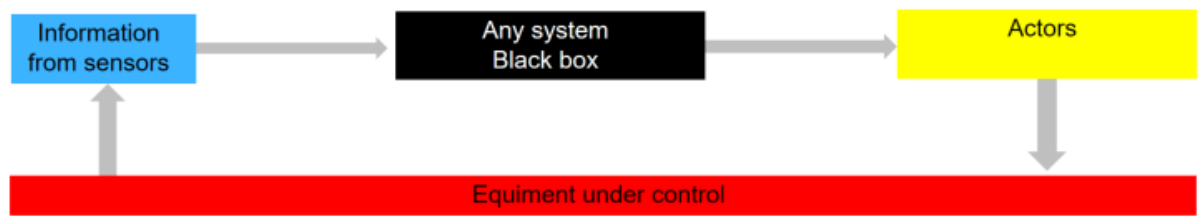

Figure 2        Arbitrary control system (black box)

Now, the black box can also be an AI system. Therefore, also a safety integrity level can be necessary if the AI system fulfills safety relevant tasks and the SIL can be determined by the same methods as for an E/E/PE system. Only the rules for the assessment of the SIL may be different depending on the type of system that implements the black box.

What SIL would we have to expect for different AI applications? This would mainly depend on the failure consequence and if other risk mitigations are possible:

- Data processing – depends on the results and what is done with it

- Assistance systems – normally no SIL if a human can always override the system

- Speech recognition – depends on what is done with the result and whether there are safe backups

- Face recognition – depends on what is done with the result, i.e. which functions are activated

- Nursing robots – giving medicine, carrying patients, so surely a SIL would be required

- Autonomous driving systems – can lead to accidents, so a SIL would be required

In any case, a hazard and risk analysis needs to be carried out to determine the SIL – or the fact that it is not necessary to determine one. The relevant functional safety standard has to be applied.

## Looking inside AI

### AI architecture

Figure 3 shows a very simple architecture of an AI system. The architecture has been inspired by Wand (2017) but does not resemble it.

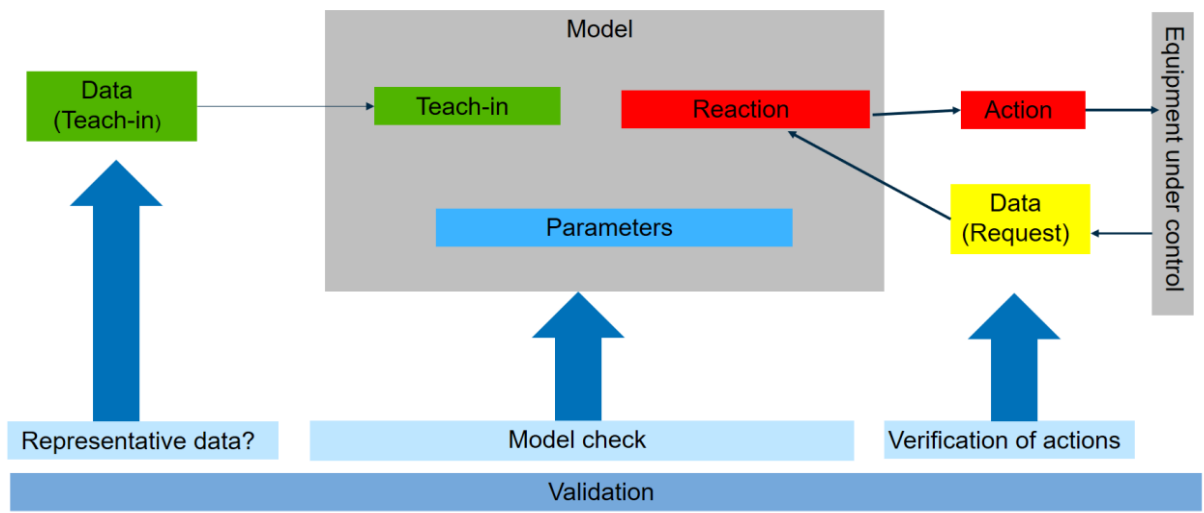

Figure 3 – Architecture of an AI system

Inside the AI system is the model, the most important feature. This model is flexible and needs to undergo a teach-in. This is done on the basis of some data. These data must be representative, i.e. they must be adequate to resemble future situations. It is necessary to avoid situations as mentioned e.g. reported by Corni (2019), where an AI system shows racism, which was imported via a non-representative set of data for learning.

After teach-in, parameters are set in the model. This is later used to generate reactions to request data and activate actors in order to control the equipment under control. Possibly, teach-in can continue even after the system has been put into exploitation.

Then it is important to

- Check the model,

- Check the representativeness of the data,

- Verify the data – model reaction – action chain, and to

- Carry out an overall validation.

In the following subsections, we will take a deeper look into several types of AI systems. This will refine the model part of the architecture described in figure 3.

## Looking at AI by Similarity Analysis

As explained by figure 3 most AI algorithms rely on or are at least similar to statistics. So as a first approach to explore the requirements for use of AI in safety applications we could what a statistical procedure would have to fulfill if we wanted to use it for safety applications This can also be interpreted as a kind of similarity analysis. What can we learn from statistical procedures? What would be the consequences if AI algorithms e. g. machine learning could just be interpreted as statistical data fitting – but with very complex algorithms and big data?

To explain the situation intuitively, let's use one of the simplest statistical models, which every engineer knows from school: linear regression i. e. fitting of a (straight) line to data. What can we learn in general from it? Note that this observation is not new, Pearl and Mackenzie already stated that neural networks "…are driven by a stream of observations to which they attempt to fit a function, in much the same way that a statistician tries to fit a line to a collection of points." But to the knowledge of the authors this similarity has not been fully exploited yet.

Let us assume that some safety-critical decision would depend on the goodness of the fitted curve. A very good example what can go wrong has been constructed by Anscombe (1973). In his data sets, see figure 4, all relevant statistical measures are equal to at least two decimal places, although obviously the sets appear very different.

Figure 4 gives some examples of a correct fit (data set 1); a data set (2), where obviously the wrong model was used; a data set (3), which is influenced by an outlier; and data set (4) with a leverage point, which results from a completely inadequate experimental design. Even from this simple example we can draw some important conclusions:

1: The model must be correct – otherwise we will never fit the data well (see data set 2), no matter how long we learn or how good the data might be.

2. The training data must be representative of the real data; particular we must make sure that the sampling is adequate (see data set 4)

3. We must have means to detect outliers (and even to remove them, see data set 3) or even Black Swans

4. We need a measure of goodness of fit (like R2 in normal regression). But such a measure and the calculated fit depends on the loss function (see data set 1, where the usual least squares loss function is assumed like in all other fits in figure 4)

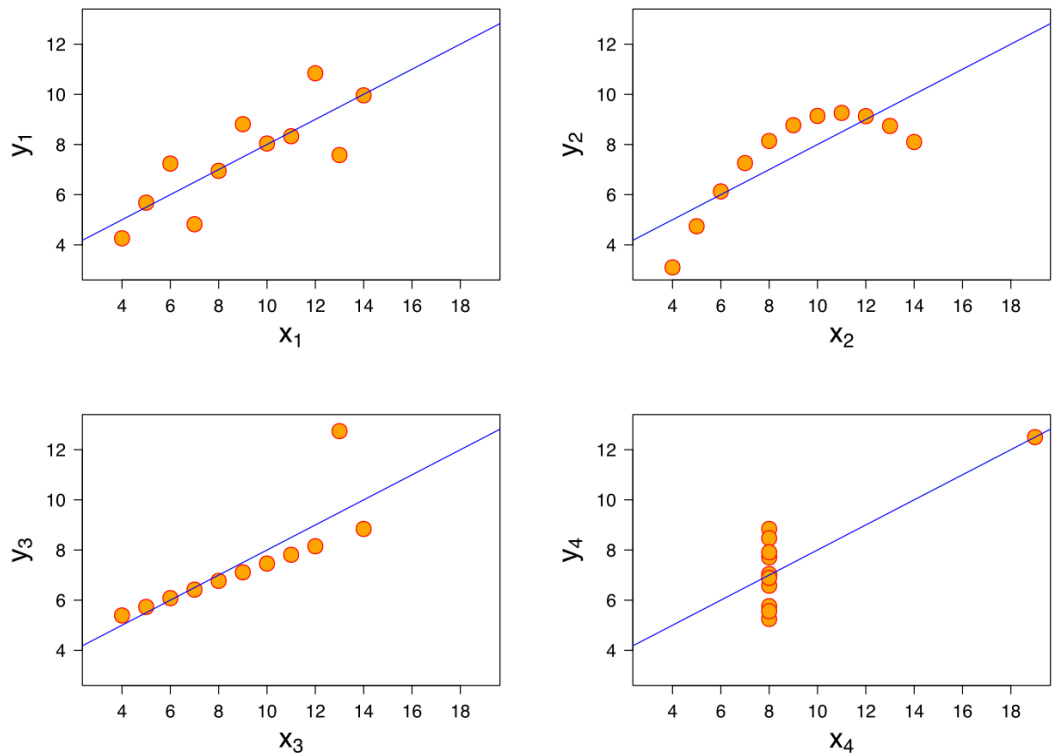

Figure 4 – Examples of what can be learned from linear regression

## Machine Learning as a classification problem

Machine learning (ML) is a particularly successful variant of AI. Statistically it can also be interpreted as a classification problem, which provides another look on the problem. So, all our findings in the preceding section directly hold for ML. Basically, most ML algorithms solve classification problems, similar to cluster or discrimination analysis in statistics. We have (at least) two classes of (big) data in a high dimensional space., see figure 5 for an illustrative two-dimensional example.

An optimal discrimination function would separate the classes completely for the training set. We may assume that a true („correct") discrimination function exists (the red curve in figure 5), but in practice ML algorithms calculate an approximation of the true function. However, there remains some space between the two classes and there exists no unique solution for the problem.

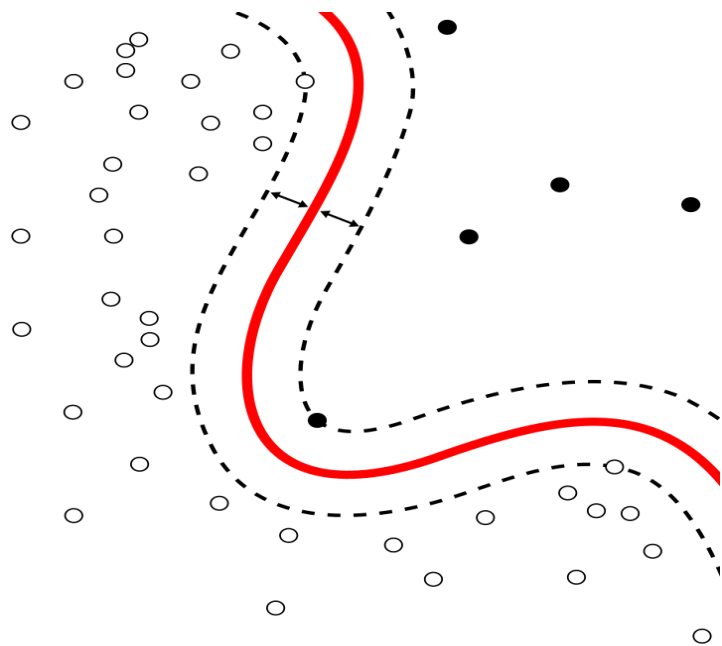

Figure 5 – Discrimination of two data sets in classification

## Artificial Neuronal Networks and the General approximation Theorem

The most polular and recently most successful variant of ML algorithms are Artificial Neural Networks (ANN). Each ANN has at least two layers that are connected by weights. A simple example is shown in figure 6.

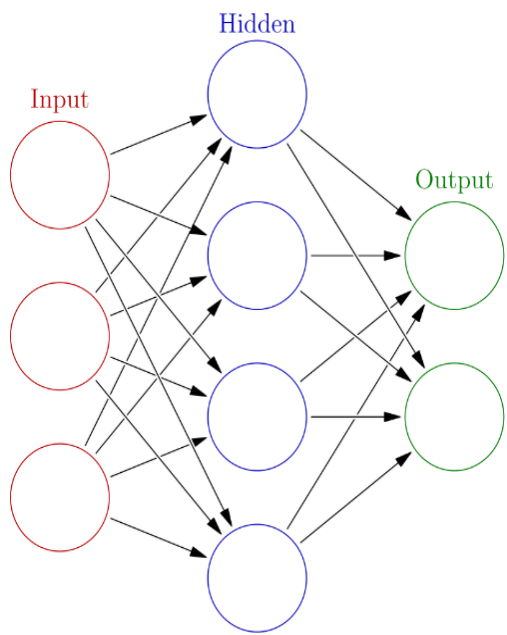

Figure 6 – artificial neural network with two layers

A mathematical model of this simple ANN can be described ed as follows: the input data vector x is transformed by weights v and w, offsets b and an output function $\phi$ (non-constant, bounded and continuous) to two output classes

$$F(x) = \sum_{i=1}^{N} v_i \, \varphi(w_i^T x + b_i) \tag{1}$$

The optimal weights for a particular cost function C, which is defined in addition to (1), are found iteratively based on the training data and a numerical algorithm.

More complex ANN add additional hidden layers (often called deep networks), but the mathematical description and solution is similar.

From our general discussion above immediately the following questions arise:

1. Is F the correct function to discriminate the data well?

2. Does it approximate the true function well?

3. Or do we need more layers or more complex functions?

4. How can we make sure that the training data are representative?

5. How can we detect outliers?

6. How can we justify the cost function C?

If we cannot answer the questions sufficiently, we might have systematic flaws in the model!

Fortunately, for question 1 there exist a variety of so called "universal approximation theorems", that show convergence of F to f, the true function, provided $\varphi$ is a bounded and continuous function and if f is continuous, see Cybenko (1989). Note that this is convergence as in the calculus definition, not some stochastic convergence.

This is quite a strong result, but it has implications related to the other questions. The most limiting assumption is the continuity of the true function f, which means that our problem space must be separable by a continuous function. And also $\varphi$ must be continuous, so we can't use jump functions for the decision making.

At first glance this result is surprising because it already holds for ANN with a single hidden layer but on second thought the results are quite obvious and a have a simple explanation:

1) F is a kind of general linear approximation to f. But it is obvious that such linear approximation for a continuous function f should be possible if the number of nodes N is sufficiently large. Also, in the classification example in figure 5 f could be approximated by stepwise linear functions.

2) Also, deep ANN with several hidden layers could be represented by single layer (with large N). Just think that the true function f would be the function represented by the multi layer network, which by the approximation theorem again could be approximated by a single layer function F.

For dependable applications, the requirements to answer question 1 could be:

1) Choose a single-layer ANN with sufficiently large N. N could be determined by a convergence criterion as known from calculus.

2) The more difficult assumption that needs to be justified would be that the data sets can be separated by a continuous function. This argument would depend on the type of application data and can hardly be general.

3)      Choose an appropriate cost function C (with justification).

## Data and Goodness of fit

The second question deals with the adequacy of the training data and also with the associated stopping rule: when is training finished?

Representative data means that teach-in must occur in a typical environment for this type of system and the environment must be such that the influences are typical for this type of use, including all the changes in the environment. So, all replications of the system (after teach-in) must be operated at least in similar environments and all replications of the system must be similar, compare Braband et al (2018). Here we must in particular also take care of the Black Swan problem (related to question 3). Possibly we have to introduce safety-related application rules for the environment in which the system will operate.

Another question is goodness of fit. How do we measure goodness-of-fit for the training data? Can we accept failure in training data? Generally, any misclassification in training data could lead to a high proportion of classification failure in practice. Take as an example the black point on the boundary line in figure 5. Assume now that both data sets are separated by the true (red) function f in figure 5. If this particular point is mis-classified, a whole set of points close to the black point would be misclassified, too, resulting in a high failure rate. On the other hand this point might also be an outlier.

This means

1.      Either we have 100% correct classification in the training data, or

2.      We can calculate the error probability well

The problem is that we cannot simply count classification errors. We have to weight them according to their importance, which may be difficult in high-dimensional spaces and big data.

Furthermore, teach-in has clearly statistical aspects. This means:

- Confidence bounds need to be taken into account.

- Derived parameters are random values containing some spread

- The subsequent decisions of the AI will also be random, with some errors:

    o   First kind error: wrong decision, although the input data are in the „right" domain

    o   Second kind error: input data are in the „wrong domain", but decision is „right".

As a consequence, the AI will have a failure probability. This must be taken into account, assigning part of the budget of the rate of dangerous failures to the AI (here: the algorithm).

## The position of functional safety standards on AI and a possible assessment procedure

If AI is used for safety relevant applications, the standards on functional safety would come into play. We consult the basic standard, IEVC 61508.Requirements of the functional safety standards – example: IEC 61508. The main information is contained in IEC 61508-3, table A.2:

> no. 5 - Artificial intelligence / fault correction SIL 2- SIL 4: NR (see C.3.12)

> no. 6 - Dynamic reconfiguration SIL2 – SIL 4: NR (see C3.13)

In part IEC 61508-7 an explanation can be found, what ai means in the terms of the standard

C.3.9 Artificial intelligence

> Fault forecasting (calculating trends), fault correction, maintenance and supervisory actions may be supported by artificial intelligence (AI) based systems in a very efficient way in diverse channels of a system, since the rules might be derived directly from the specifications and checked against these. Certain common faults which are introduced into specifications, by implicitly already having some design and implementation rules in mind, may be avoided effectively by this approach, especially when applying a combination of models and methods in a functional or descriptive manner. The methods are selected in such a way that faults may be corrected, and the effects of failures be minimised, in order to meet the desired safety integrity.

In fact, the IEC 61508 sees AI as a means for fault correction and dynamic reconfiguration as a reaction of a fault in the control system. Such an application would make the control system unpredictable.

How to cope with the IEC 61508 rules against artificial intelligence? The statement in the standard is combined with a statement about dynamic reconfiguration, which is an undesired for SIL 2 …SIL 4. If AI is implemented in the control system itself, this would not be a reaction on faults of the control system, it would be a feature.

The functional safety standard requires a predictable system. Predictable means that measures against systematic failures so that they can be neglected. Random failures' occurrence is brought to a sufficiently low level.

Therefore, AI system's behavior must be predictable in a statistical sense. Note that this predictive behavior here is not a deterministic behavior, but a statistically predictable behavior. This means that the AI system will contribute to random dangerous failures that would be caused by a random behavior of the software itself. This is a key difference to normal E/E/PE systems, where software is considered deterministic with systematic errors only requirements and following the software requirements of the functional safety standards would reduce them to an acceptable level.

An assessment approach can then be based on the following steps:

- Analyzing the model,

- Taking part of the budget for random failures for the AI system since it shows probabilistic behavior,

- Treat the AI system as a normal mathematical model, but only with probabilistic behavior.

Then assessment is carried out in the same manner as a normal safety assessment with a complicated mathematical model. It is not the intention of the author to repeat the procedures of safety assessment. For details of an assessment process see e.g. Wigger (2018).

The main part of the assessment is the model check.

The mathematical model needs to be checked regarding the following aspects:

- correctness of the model according to physical / chemical / mathematical and other scientific proven theories,

- equivalence to other mathematical models as e.g. of brake curves, thermal models etc.

That means, the theory / model must be disclosed to the assessor. The models might be of one of the following types, see e.g. Wang (2018):

- Neural network,

- Long short-term memory,

- Auto encoder,

- Deep Boltzman machine,

- Generative adversarial network,

- Attention-based LSTM.

The more flexible the model, the more complicated its analysis will be. In the next section we provide an example on how such a model analysis could be carried out for a very simple model.

The great effort for model checking leads to the question, whether proven in use approaches could be applied. According to Braband et al (2018) this would mean to accumulated a minimum number of failure free hours according to the following scheme:

- $3 \ 10^6$ failure free hours for SIL 1

- $3 \ 10^8$ failure free hours for SIL 4

Practical experience shows that it is hard to accumulate such a quantity of failure free hours. As a result, model analysis as one of the main parts of safety assessment needs to be done.

## Academic Example

Assume a classification system that classifies objects in two categories: „left" and „right" based on one real-valued parameter. The parameter is assumed to be normally distributed. Note that statistically the model is completely defined by this assumption, which would have to be justified in practical applications. It can't be taken for granted, and for this reason we label it as an academic example as we assume to know the true model.

There are two sub-populations characterized by the following distributions:

- „left" is characterized by a normal distribution with mean $m_L$ and spread $\sigma_L$,

- „right" is characterized by a normal distribution with mean $m_R$ and spread $\sigma_R$.

First, assume the parameters to be known.

Then the following classification rule is established:

„left" if X≤z and „right" if X>z,

where is a „properly" chosen constant. Now the first kind error and the second kind error can be computed

$\alpha = 1 - \Phi(z-m_L/\sigma_L)$ \qquad first kind error, \qquad (2)

$\beta = \Phi(z-m_R/\sigma_R)$ \qquad second kind error, \qquad (3)

$\Phi(z-mL/\sigma_L)$ \qquad correct „left" classification, \qquad (4)

$1- \Phi(z-m_R/\sigma_R)$ \qquad correct „right" classification, \qquad (5)

$\Phi$ – standard normal integral.

The first kind error is the probability that an object is classified in the sub-population "right" although it belongs to "left". The second kind error is the probability that that an object is classified in the sub-population "left" although it belongs to "right". The parameters $\sigma_R$ and $\sigma_L$ should be as small as possible to have small errors.

Now there is one missing point. Parameters $m_L$, $m_R$, $\sigma_L$ and $\sigma_R$ are not known but must be obtained by a statistical procedure that means that they must be learned from a sample of data.

How does the system learn? The system learns from two samples for the both sub-populations:

A „left" sample $XL_i$, i=1,$n_L$ and a „right" sample $XR_i$, i=1,…,$n_R$ are used for teaching.

From the samples, the unknown parameters can be estimated:

$m_R = (1/n_L) \Sigma XR_i$, \qquad (6)

$m_L = (1/n_L) \Sigma XL_i$, \qquad (7)

$\sigma_R^2 = \Sigma (XR_i - m_R)^2/(n_R-1)$, \qquad (8)

$\sigma_L^2 = \Sigma (XL_i - m_L)^2/(n_R-1)$. \qquad (9)

The point estimators of statistical characteristics are given in italics. The sum runs over the index i for 1 to $n_L$ or $n_R$, respectively.

In a next step the confidence limits for the parameters have to be used instead of the point estimators given by (6) – (9). Confidence limits will be chosen as such that the misclassification error becomes large, i.e. upper bounds for the sigmas and $m_L$ and a lower bound for $m_R$. We use single parameter bounds – not combined ones - to simplify the computation.

The point estimators (6) – (9) have the following distributions:

$(n_L-1) \sigma_L{}^2 / \sigma_L{}^2$ is chi-squared distributed with $n_L-1$ degrees of freedom

$(n_R-1) \sigma_R{}^2 / \sigma_R{}^2$ is chi-squared distributed with $n_R-1$ degrees of freedom

$\sqrt{n_L}(m_L\text{-}m_L)/\sigma_L$ has a t distribution with $n_L-1$ degrees of freedom

$\sqrt{n_R}(m_R\text{-}m_R)/\sigma_R$ has a t distribution with $n_R-1$ degrees of freedom

The least favorable values are:

upper confidence bounds for the variances, i.e.

$$\sqrt{(n_R - 1)/Chi2(n_R - 1; 1 - \gamma)}\,\sigma_R, \tag{10}$$

$$\sqrt{(n_L - 1)/Chi2(n_L - 1; 1 - \gamma)}\,\sigma_L \tag{11}$$

where Chi2(n;1-$\gamma$) is the quantile of the Chi-squared distribution with 1-$\gamma$ coverage and

the lower confidence bound for $m_L$

$$m_L\text{-}t(n_L\text{-}1;\gamma)\sigma_L/\sqrt{n_L} \tag{12}$$

and the upper confidence bound for $m_R$

$$m_R\text{+}t(n_R\text{-}1;\gamma)\sigma_R/\sqrt{n_R}, \tag{13}$$

where t(n;$\gamma$) is the quantile of the t distribution with n degrees of freedom and coverage 1-$\gamma$.

Inserting the confidence bounds (10) – (13) into the formulae (2) – (5) gives the probabilities of errors.

If misclassification with a type one error is dangerous, (1) with (6) and (8) gives the probability of a dangerous failure. However, to account for errors coming from the confidence intervals, value

$\alpha+2\gamma$

Released

should be used. The interpretation of $\gamma$ as a probability that the true value lies outside the confidence interval is not a frequentist one, but a Bayesian using an appropriate prior.

For a SIL 1 system, a probability of failure on demand of 0.1 must not be exceeded. This value can be seen as a budget:

One might give 0.05 as a maximal value for hardware failures and 0.05 for the AI algorithm. The latter can be split according to

$0.05 = \alpha + 2\gamma$

e.g. in the form

$\alpha = 0.025, \gamma = 0.0125.$

For a SIL 4, IEC 61508 provides a threshold value of 0.0001 for the probability of failure on demand.

The reader might repeat the calculation. As a further exercise, she might consider conditions on m and the Sigma values to fulfil the requirements. This simple example shows that complicated computations are to be expected. Even with this very simple example, we were confronted with complex mathematics.

What is now the way out of this complicated situation?

There exist mainly two options:

1. The AI system does not need a SIL since its behavior does not have critical consequences (no injuries to persons etc.)

2. The AI system is supported by a sufficiently simple E/E/PE system, having the necessary SIL, that checks all dangerous decisions according to simpler algorithms and inhibits dangerous reactions

The options need to be supported by a risk analysis (see IEC 61508).

## Research Challenge

We admit that the example is quite simple and academic, but we believe that we need to understand and solve small problems first before we can approach high-dimensional problems.

In order to take a little bit more practical example, consider the following problem: You are given a set of n two-dimensional points which are classified into two sets (like figure 5, but only the points). The model is unknown, but you can control the number of points to a certain extent. You do not know anything else but that the decision problem is safety-related with SIL x. You may choose your favorite classification method, e.g. ANN.

Under which assumptions can you provide a safety argument according to an acknowledged safety standard e. g. IEC 61508? Can you also provide reasonable guidance how the validity of your assumptions may be checked in practice?

This may seem a simple problem, but it has high leverage: If we can't provide a safety argument (under assumptions that can reasonably be checked in practice) then (at least some classes of) AI algorithms can't be used for safety-related applications. But if we can solve the problems under certain conditions, we might be able to generalize the approach to higher dimensions.

# Conclusions

In this paper we have described a possible approach to safety assessment of AI systems although several questions remain open and may only be solved in the context of a particular application.

A Safety Integrity Level can be determined as for a normal E/E/PE system. This has to be substantiated by a hazard and risk analysis. This is also necessary, if the system does not require a SIL.

AI can be easily used in situations, where no critical consequences occur, which has to be supported by a risk analysis. Then, no safety integrity level requirements need to be implemented in the system and safety assessment is not necessary.

We have proposed an approach to analyze the model. The analysis to be carried out depends very much on the type of model. An assessment requires always an in-depth model analysis of the model of AI, that means AI as such cannot be analyzed since it covers a lot of different approaches. The more flexible the model, the more complicated the analysis has to be. For use in critical systems it seems a useful approach is to restrict the type of models in order to simplify the design and the assessment of the AI system.

Pearl and Mackenzie (2018) have approached the problem from a similar angle and have concluded that causality needs to be introduced into AI, before we can rely on its conclusions. One of their conclusions is that it is necessary to "formulate a model of the process that generates the data, or at least some aspects of that process".

We have provided an academic example in order to show how one would have to proceed for this specific type of model.

Finally, we have introduced a research challenge whose solution might be decisive for the use of AI algorithms for safety-related applications. The challenge is to formulate a model of the data generation process that allows a safety analysis and that can be justified to hold in practical applications.

In order to use AI systems without the burden of an extensive safety assessment there are only two possibilities: either have an AI system that is not safety relevant or have another safety relevant E/E/PE system that take over full responsibility for safety.